# Oblique Stripe Removal in Remote Sensing Images via Oriented Variation

Xinxin Liu, Xiliang Lu, Huanfeng Shen, *Senior Member*, *IEEE*, Qiangqiang Yuan, *Member*, *IEEE*, Liangpei Zhang, *Senior Member*, *IEEE*

***Abstract*—Destriping is a classical problem in remote sensing image processing. Although considerable effort has been made to remove stripes, few of the existing methods can eliminate stripe noise with arbitrary orientations. This situation makes the removal of oblique stripes in the higher-level remote sensing products become an unfinished and urgent issue. To overcome the challenging problem, we propose a novel destriping model which is self-adjusted to different orientations of stripe noise. First of all, the oriented variation model is designed to accomplish the stripe orientation approximation. In this model, the stripe direction is automatically estimated and then imbedded into the constraint term to depict the along-stripe smoothness of the stripe component. Mainly based on the oriented variation model, a whole destriping framework is proposed by jointly employing an $\ell^1$-norm constraint and a TV regularization to separately capture the global distribution property of stripe component and the piecewise smoothness of the clean image. The qualitative and quantitative experimental results of both orientation and destriping aspects confirm the effectiveness and stability of the proposed method.**

***Index Terms*—Destriping, oblique stripes, optimization-based model, oriented variation, remote sensing image, stripe orientation.**

## I. Introduction

REMOTE sensing images acquired by multiple-detector imaging systems often suffer from stripe noise. This degradation problem, which is mainly due to the inconsistent responses to perceived energy among the different sensors, can badly affect the data quality and further limit the precision in the subsequent remote sensing applications [1], [2]. To alleviate the striping effects, remote sensing images are generally experienced preprocessing procedure before providing to users. However, this preprocessing only plays a fairly limited role due to the imperfect calibration coefficients (computed either through the uniform light source in the laboratory or onboard the satellite) or the variability in the sensor responses [3]. Therefore, image-based destriping is both necessary and important.

In the past few decades, a number of image-based destriping methods have been proposed. The first family of destriping techniques focus on ensuring and truncating the stripe-related component in a transformed domain. Typical examples include the Fourier domain filter [1], [4], wavelet analysis [5], [6], and the combined wavelet-Fourier filter [7], [8]. The filtering-based models are able to better process stripes with a relatively concentrated frequency, such as periodic stripes, and even in geo-rectified images, they are still usable. However, if some healthy details possess the same frequency as the stripes, more information than the stripe noise will be simultaneously filtered out. Although the filtering-based methods are easy and efficient to implement, the blurring or ringing artifacts introduced by these methods inevitably damage the accuracy of the data.

Another popular group of destriping methods, which include the equalization [9], histogram modification [10], [11], and moment matching [12] methods, rely on the statistical property of the digital numbers for each sensor. Given that different sensors record statistically similar sub-images [12], the statistical-based approaches then remove the stripes by adjusting the target sensor distribution to a reference one. However, since the similarity assumption does not always apply to different images, especially when the images are covered by complicated or heterogeneous background information with a relatively small image size, the destriping performance can sometimes be highly limited. To overcome this limitation, the authors in [13] used an improved technique combining with histogram matching and a facet filter to tackle noisy stripes in Moderate Resolution Imaging Spectroradiometer (MODIS) data, and in [14], a piecewise method using local statistics was proposed.

Until recently, scholars have shown considerable interest in optimization-based models. By designing appropriate constraint terms in the destriping model using the stripe or image features, the desired stripe-eliminated results can be obtained via minimizing the energy function. Under a maximum *a posteriori* framework (MAP), Shen and Zhang [15] provided a model for both destriping and inpainting

This work was supported in part by the National Natural Science Foundation of China under Grants 61671334 and 41661134015 and in part by the Fundamental Research Funds for the Central Universities.

Xinxin Liu is with the College of Electrical and Information Engineering, Hunan University, Changsha 410082, China (e-mail: liuxinxin@hnu.edu.cn).

Xiliang Lu is with the School of Mathematics and Statistics, and with the Hubei Key Laboratory of Computational Science, Wuhan University, Wuhan 430072, China (e-mail: xllv.math@whu.edu.cn).

Huanfeng Shen is with the School of Resource and Environmental Sciences, and with the Collaborative Innovation Center of Geospatial Technology, Wuhan University, Wuhan 430079, China (e-mail: shenhf@whu.edu.cn).

Qiangqiang Yuan is with the School of Geodesy and Geomatics, Wuhan University, Wuhan 430079, China (e-mail: yqiang86@gmail.com).

Liangpei Zhang is with the State Key Laboratory of Information Engineering in Surveying, Mapping, and Remote Sensing, and with the Collaborative Innovation Center of Geospatial Technology, Wuhan University, Wuhan 430079, China (e-mail: zlp62@whu.edu.cn).

problems through the help of a Huber-Markov prior. Bouali and Ladjal [16] used the structural information of stripe noise and proposed a more sophisticated unidirectional total variation (UTV) model to treat stripes distinctively in the along-stripe and across-stripe directions. Liu *et al*. [17], from a different angle, designed a stripe separation model according to the local and global properties of stripe noise. Aiming at suppressing stripe noise in hyperspectral data, Chang *et al*. [18] employed a spectral-spatial TV model, while Lu *et al*. [19] and Zhang *et al*. [20] utilized low-rank representation to separately process destriping and mixed noise removal problems.

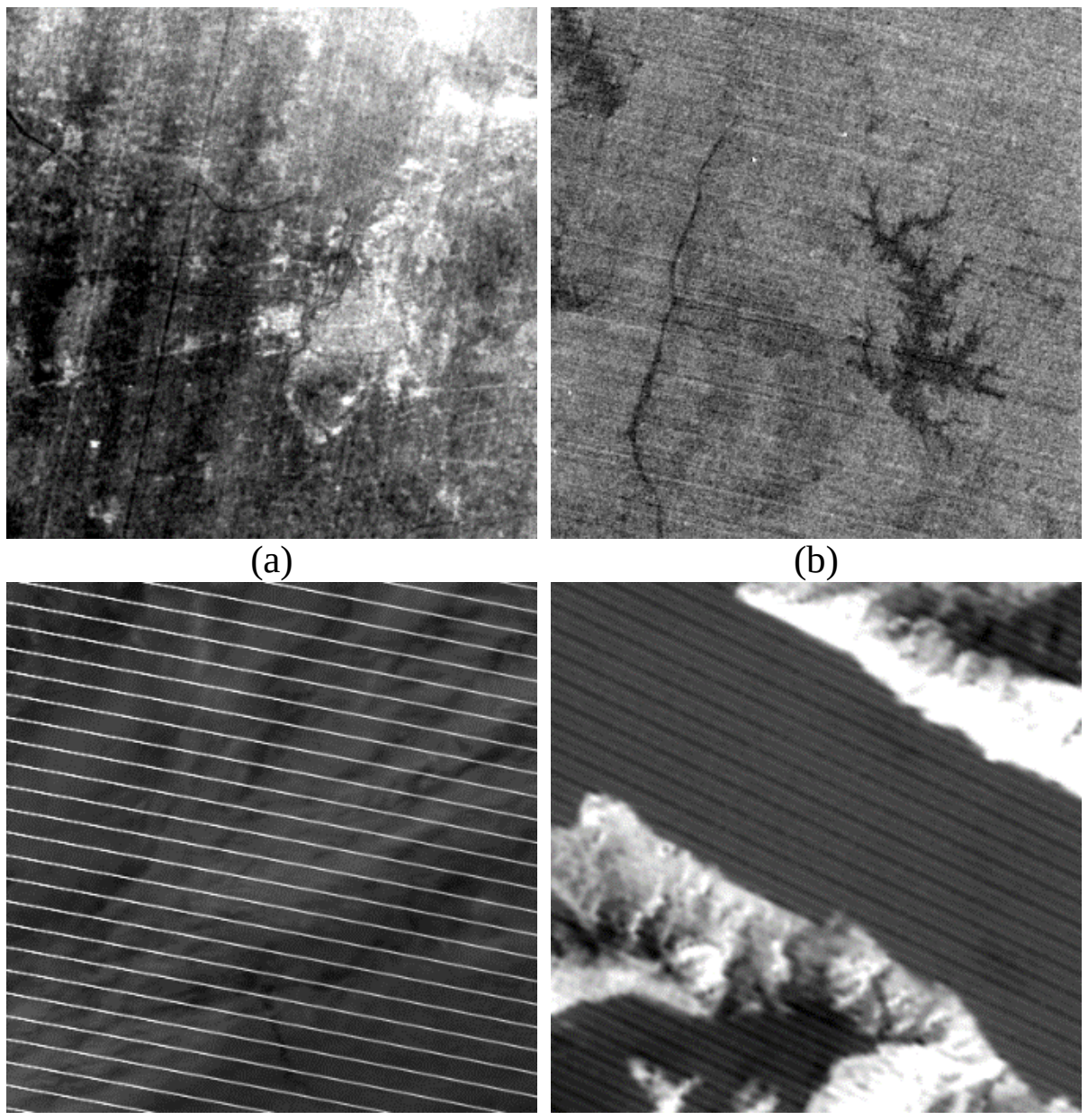

(a) (b) (c) (d)

Fig. 1. Several examples of oblique stripes in real remote sensing data: (a) HJ1A/HIS data; (b) Sentinel 2A data; (c) Terra MODIS data; and (d) LANDSAT 5 TM thermal data.

Unfortunately, most of the existing image-based destriping methods only focus on processing the horizontal and vertical stripes, while the removal of oblique stripes is still a problem. Stripes acquired by the sensors are either horizontal or vertical in the original remote sensing images, but in the process of producing higher-level remote sensing products, the geometric registration will transform those stripes into the oblique ones. Since higher-level remote sensing products often possess more value and utility in the scientific application, oblique stripe issue needs to be solved. This issue becomes even urgent when the officially released products of a certain satellite platform contain only geo-rectified images, e.g., HJ1A and Sentinel 2A. Several examples of oblique stripes in real remote sensing data are shown in Fig. 1. To address the oblique stripe issue, an intuitive approach is to convert the oblique stripes into the horizontal or vertical direction through a specific-angle image rotation; however, the resampling step after the image rotation inevitably introduces new errors into all the pixels [21], such as transforming the original line pattern of the stripe noise into a jagged one. Although the filtering-based models show potential for oblique stripes, their blurring or ringing artifacts are always barely controlled when processing stripes without periodicity [12], [13], [18].

In the present paper, the oriented variation model is first proposed to approximate the stripe orientation (estimated in Fourier domain) through describing its along-stripe smoothness. Then, an optimization-based framework is designed from an image decomposition perspective [22], [23] to eliminate oblique stripes via combining the oriented variation model and another two constraints on global distribution property of stripe component ( $\ell^1$ -norm regularization) and the piecewise smoothness of the clean image (TV techniques). Numerous images were tested in the experiments to verify the effectiveness and stability of the proposed orientation and destriping methods.

The remainder of this paper is organized as follows. Section II and Section III introduce the proposed oriented variation model and the destriping framework, respectively. Section IV describes the large number of experiments undertaken in terms of both the orientation and destriping aspects to demonstrate the performance of the proposed approach. A further discussion is provided in Section V, and our conclusions are drawn in Section VI.

## II. Oriented Variation Model

Unlike other kinds of noise, the structural characteristic of stripe noise is quite remarkable and unique, and is embodied as a parallel texture with a specific orientation. The oriented property of stripe noise can be regarded as the clear smoothness in the along-stripe direction with small differences between adjacent pixels inside each single stripe. Based on this fact, the minimization of the along-stripe variations can thus be easily achieved via unidirectional gradient techniques [16], [18] for horizontal and vertical stripes. However, when taking oblique stripes into account, their orientations complicate the practical process. An intuitive solution is to rotate the oblique stripes into the horizontal or vertical direction before conducting destriping. This strategy seems reasonable, but brings a new problem as the image resampling generates jagged effects which severely damage the line pattern (along-stripe variational property) of stripes. An illustration is given in Fig. 2. Therefore, a direct destriping manner was finally chosen for the oblique stripe removal task in the present work.

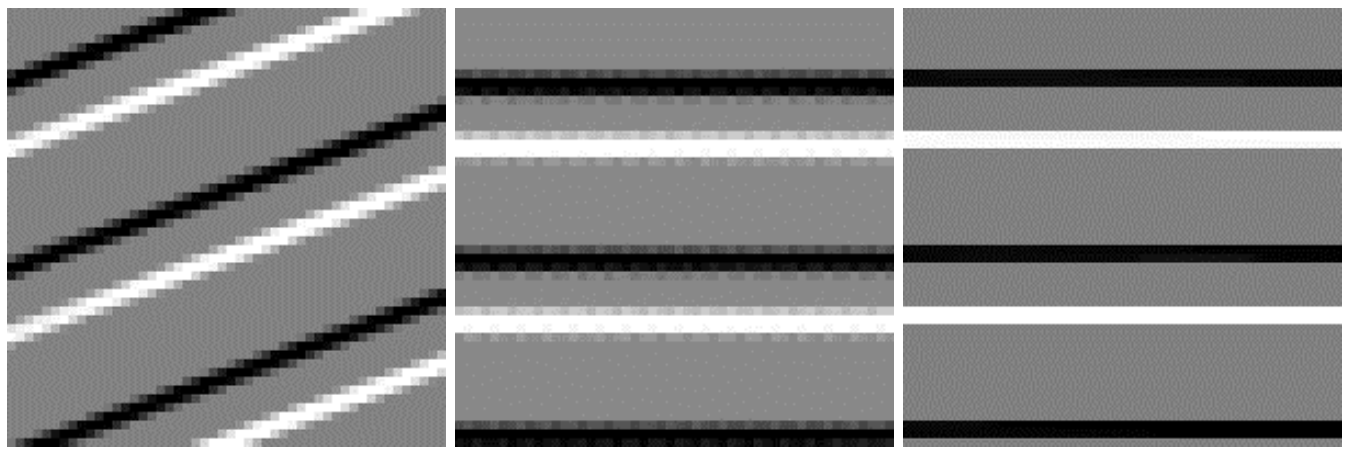

(a) (b) (c)

Fig. 2. Illustration of jagged stripes after image rotation: (a) oblique stripes; (b) jagged stripes caused by rotating (a) into horizontal direction; and (c) normal horizontal stripes.

Generally speaking, stripes come with an invariant horizontal or vertical direction in the original remote sensing images. Even after geo-rectification, the direction of oblique stripes can still be thought of as invariant in a certain size of image. Thus, the inherent orientation of stripe noise is relatively stable. To take advantage of this implicit information to capture the explicit parallel pattern of oblique stripes, orientation for oblique stripes is crucial.

Under the assumption that the stripe direction is steady in a certain image, an orientation strategy from a global point of view is much easier and more conforming to capture the stripe direction than those from a local perspective. For instance, many multiscale geometric analyses including wavelet transform feature a localized analysis, but the detailed direction of each local region is always affected by the natural image textures and cannot well reflect the stripe orientation. Therefore, two-dimensional fast Fourier transform (FFT) that obtains the structure direction by virtue of the global frequency analysis in the Fourier domain is thoughtfully chosen in this work due to its simpleness and suitability.

As an effective tool, Fourier transform has been successfully exploited in many orientation-related tasks, such as image rotation [24], image registration [25], and some other applications. However, due to the mixture of the stripe pattern and background information in the observed data, it is always difficult to calculate precise stripe orientation via direct FFT on a noisy observation [see Fig. 3(a) and (b)]. Inspired by the texture decomposition strategy [22], [23], [26], a background information removal process is considered before FFT to enhance the dominance of the stripe noise for a more accurate orientation. The guided filter [27], which is known for its structure-transferring property and efficient computation, is then implemented on the observed image to realize the elimination of the background information through the following approach:

$$\mathbf{E} = t\left(\mathbf{Y} - G(\mathbf{Y},\mathbf{Y})\right) \tag{1}$$

where $\mathbf{Y}$ and $\mathbf{E}$ represent the noisy observation and the background-eliminated image; $t$ is a positive factor; and the function $G(\mathbf{Y},\mathbf{Y})$ denotes the guided filter, in which both the input and the guidance image are $\mathbf{Y}$.

Owing to the edge-preserving smoothing property of the guided filter, the smoothed output $\mathbf{B} = G(\mathbf{Y},\mathbf{Y})$ can be seen as a base layer which holds the background information of $\mathbf{Y}$. Therefore, the difference ($\mathbf{E}$, after $t$ times enhancement) between $\mathbf{Y}$ and $\mathbf{B}$ is essentially the background-eliminated image [27], [28], and its main content is those details relevant to the natural image textures and the unnatural "stripe edges". Compared to those oscillating natural textures in image $\mathbf{E}$, the oblique stripes possess a parallel structure and an almost invariant orientation, which can yield a concentrated and dominant frequency in the Fourier space. To capture this stripe-related frequency for the orientation, a simple method after a global Fourier transform of $\mathbf{E}$ is used as follows:

$$\xi = \max \mathcal{F}(\mathbf{E}) \tag{2}$$

where $\xi$ is the dominant frequency corresponding to the parallel "stripe edges" in image $\mathbf{E}$; $\mathcal{F}$ represents the Fourier transform; and $\max$ is the operation to compute the maximum frequency, except for the direct current component (DC-component) in $\mathcal{F}(\mathbf{E})$. According to the property of the frequencies in the two-dimensional Fourier domain (after centering the DC-component), the location of each frequency hints at a continuous change along the direction to the Fourier center (DC-component), but constant in its perpendicular direction [7], [29]. Given that the direction of each frequency is exactly the direction to the Fourier center, the perpendicular direction to $\xi$ can therefore be used to orient the structural oblique stripes as stripes displaying apparent smoothness (closely related to constant) in their along-stripe direction [17]. For the purpose of calculation, the stripe orientation $\theta_{stripe} (0° \le \theta_{stripe} < 180°)$ is defined as the angle rotated from the positive y-axis to the "stripe edges". The principle of oblique stripe orientation is illustrated and summarized in Fig. 3.

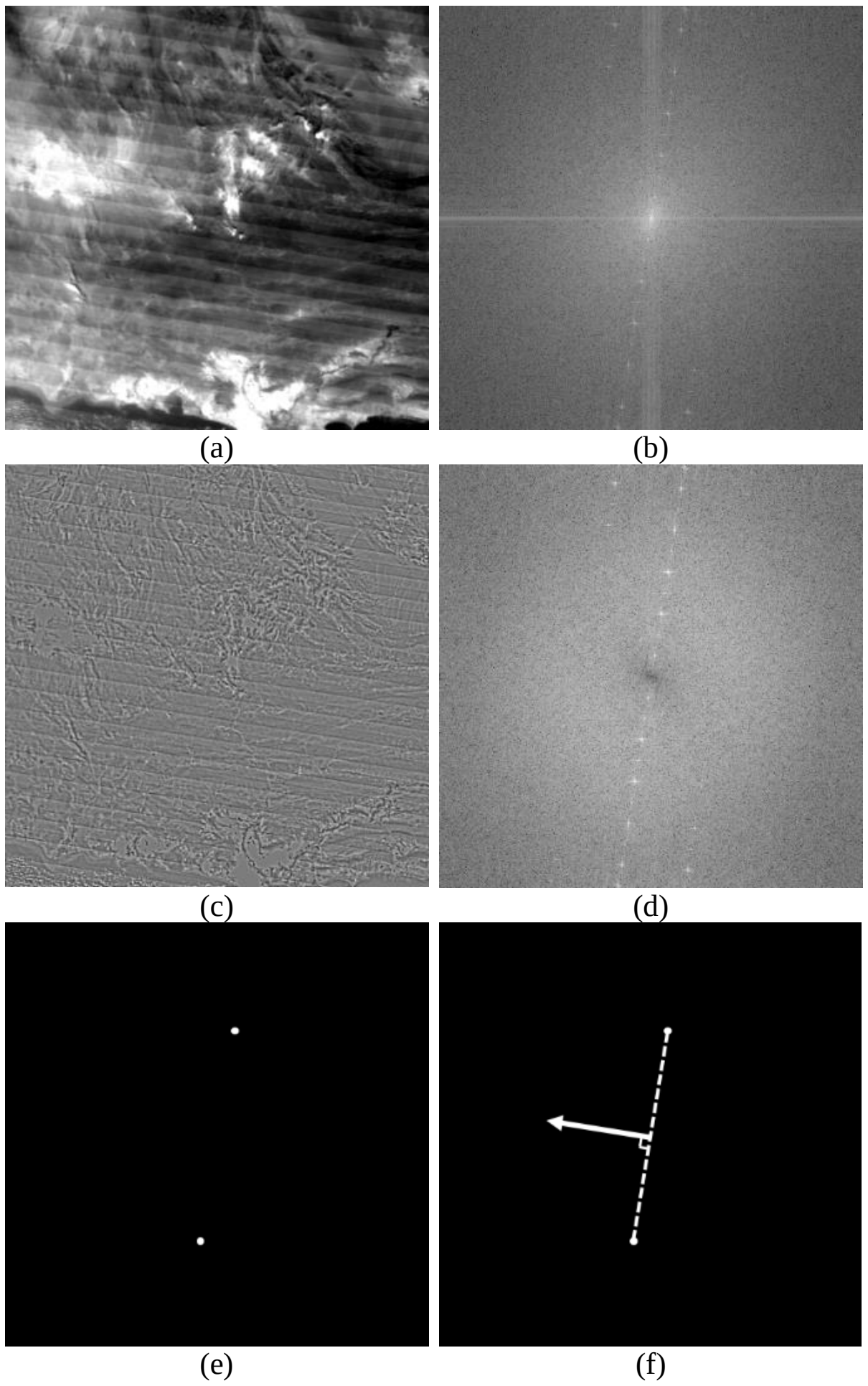

Fig. 3. Illustration of the orientation estimation for oblique stripes: (a) the input image $\mathbf{Y}$; (b) the Fourier space of $\mathbf{Y}$; (c) the background-eliminated image $\mathbf{E}$; (d) the Fourier space of $\mathbf{E}$; (e) the dominant frequency $\xi$; and (f) the estimated stripe orientation.

To imbed the estimated orientation information of stripe noise into the model, while also considering the raster property of remote sensing imagery, the concept of oriented variation is proposed referring to the direction-adaptive TV regularization applied in video compression [30], image denoising [31], and latent fingerprint segmentation [32]. The basic idea of oriented variation is to choose the candidate with the closest direction to the real oblique stripe orientation from all the candidate discrete directional derivatives to accomplish the orientation approximation. Through the aid of a certain template with the size of $(2r+1)\times(2r+1)$, centered at $(i,j)$, all the candidate directional derivatives, along with their orientations, can be explicitly defined as follows:

$$D_\theta \mathbf{S}(i,j)=\mathbf{S}(i,j)-\mathbf{S}(i+a,j+b) \tag{3}$$

where

$$\theta=\begin{cases}\arctan\dfrac{a}{b} & b<0\\ 90^\circ & b=0\\ \arctan\dfrac{a}{b}+180^\circ & b>0\end{cases} \tag{4}$$

where $a\in[-r,0]$, $b\in[-r,r]$, $\mathbf{S}$ represents the stripe components, and $D_\theta$ denotes a set of oriented differential operators matching the candidate direction $\theta(0^\circ\le\theta<180^\circ)$. For a clear explanation, we take $r=2$ as an example to detail the candidate calculation for the proposed oriented variation in Eq. (5), and we illustrate the process in Fig. 4.

$$\begin{cases}D_{\theta_1}S(i,j)=S(i,j)-S(i-1,j) & \theta_1=0^\circ\\ D_{\theta_2}S(i,j)=S(i,j)-S(i-2,j-1) & \theta_2=26.6^\circ\\ D_{\theta_3}S(i,j)=S(i,j)-S(i-1,j-1) & \theta_3=45^\circ\\ D_{\theta_4}S(i,j)=S(i,j)-S(i-1,j-2) & \theta_4=63.4^\circ\\ D_{\theta_5}S(i,j)=S(i,j)-S(i,j-1) & \theta_5=90^\circ\\ D_{\theta_6}S(i,j)=S(i,j)-S(i+1,j-2) & \theta_6=116.6^\circ\\ D_{\theta_7}S(i,j)=S(i,j)-S(i+1,j-1) & \theta_7=135^\circ\\ D_{\theta_8}S(i,j)=S(i,j)-S(i+2,j-1) & \theta_8=153.4^\circ\end{cases} \tag{5}$$

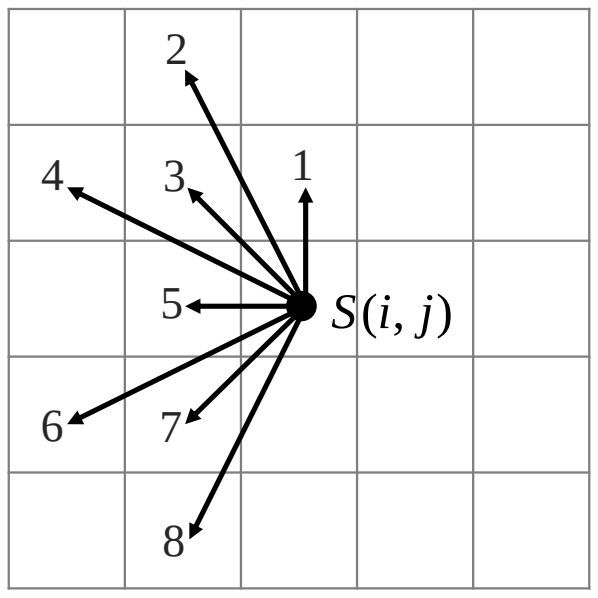

Fig. 4. Candidate directions for oriented variation in the example case.

After setting an appropriate value of $r$ to ensure an adequate number of candidate directions in the real implementation, we choose the direction closest to the real stripe orientation (estimated in the Fourier domain) from all the candidates, and use its oriented differential operator to approximate the stripe orientation through constructing the along-stripe smoothness as:

$$R_1(\mathbf{S})=OV_{\hat\theta}(\mathbf{S}) \tag{6}$$

where

$$OV_{\hat\theta}(\mathbf{S})=\|D_{\hat\theta}\mathbf{S}\|_1 \quad \text{s.t.} \quad \hat\theta=\min\theta-\theta_{stripe} \tag{7}$$

where $D_{\hat\theta}$ is the chosen oriented differential operator. Since for any value of $r$, two candidate directions $0^\circ$ and $90^\circ$ always exist, horizontal or vertical stripes are never a problem for the proposed model. Additionally, although some other methods can approximate the oblique orientation via reweighted horizontal and vertical derivatives [31], [32], the directional derivatives employed in the proposed approach give a more intuitive description for regularizing the along-stripe smoothness.

## III. Destriping Framework

### *A. Problem Formulation*

Given that oblique stripes in remote sensing images are additive noise, then the noisy observation degraded from a clean true image can be modeled as:

$$\mathbf{Y}=\mathbf{X}+\mathbf{S} \tag{8}$$

where $\mathbf{X}$ is the clean true image.

### *B. Oblique Stripe Removal Algorithm*

Oriented variation is an effective tool to describe the orientation information [along-stripe smoothness, constructed in model (7)] of oblique stripes, but is apparently not enough for constructing a whole destriping framework. Therefore, besides the feature in the along-stripe variational domain, other inherent properties of the stripe component $\mathbf{S}$ and the latent clean image $\mathbf{X}$ should also be considered. In this paper, an oriented variation model and an $\ell^1$-norm term are constrained on $\mathbf{S}$ to capture the along-stripe smoothness and the global distribution property of stripe component, while TV regularization is utilized on $\mathbf{X}$ to describe its piecewise smoothness and simultaneously maintain the sharp discontinuous edges. The flowchart of the whole proposed procedure is illustrated in Fig. 5.

*1) Stripe Distribution Modeling:* Owing to the across-stripe discontinuity of stripe noise (opposite to the along-stripe smoothness), the "edges" of oblique stripes are non-smooth. Moreover, the stripe noise does not conform to a law of

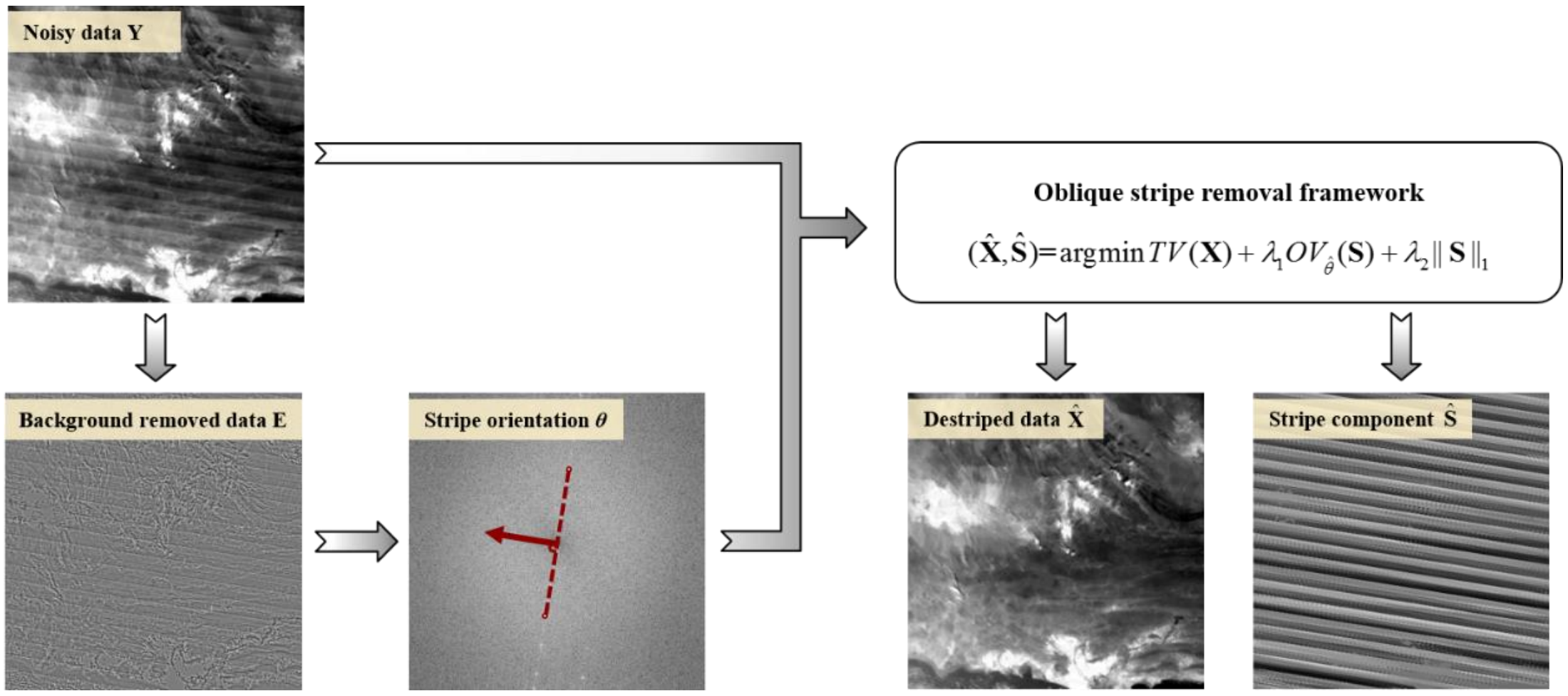

Fig. 5. The flowchart of the proposed destriping framework.

Gaussian distribution, and is not restrained to be of zero mean. Therefore, an $\ell^1$-norm is more suitable than the commonly used $\ell^2$-norm to depict the global stripe distribution, as formulated in Eq. (9). Similar conclusions can be drawn through detailed comparison between $\ell^1$-norm and $\ell^2$-norm regularization introduced in [33], [34]. It is worth mentioning that since $\ell^1$-norm is supposed to lead to sparsity, when stripes are relatively sparse in the remote sensing images, the proposed $R_2(\mathbf{S})$ will fit the case better.

$$R_2(\mathbf{S}) = \|\mathbf{S}\|_1 \tag{9}$$

*2) Clean Image Modeling:* The TV model was originally used for the denoising purpose by Rudin, Osher, and Fatemi (ROF) [35], and has since been widely exploited in a multitude of imaging problems [15], [36], [37] as an effective image prior. By minimizing the horizontal and vertical derivatives in the desired image ($\mathbf{X}$), the TV model can depict the piecewise smoothness of an image without affecting the sharp discontinuous edge information. Due to its superior edge-preserving property and its robustness to different observations, the TV model is utilized in this work to constrain the clean image $\mathbf{X}$. The formulation is given as follows:

$$J(\mathbf{X}) = TV(\mathbf{X}) \tag{10}$$

where

$$TV(\mathbf{X}) = \sum_{i,j}\sqrt{D_h\mathbf{X}(i,j)^2 + D_v\mathbf{X}(i,j)^2} \tag{11}$$

where $D_h$ and $D_v$ denote the partial differential operator in the horizontal and vertical directions, respectively.

By combining the constraint terms both on oblique stripes and the clean image, a destriping framework from an image decomposition perspective can be built as:

$$(\hat{\mathbf{X}},\hat{\mathbf{S}}) = \arg\min J(\mathbf{X}) + \lambda_1 R_1(\mathbf{S}) + \lambda_2 R_2(\mathbf{S}) \tag{12}$$

where $\lambda_1$ and $\lambda_2$ are two positive regularization parameters used to balance the functioning degree between $J(\mathbf{X})$, $R_1(\mathbf{S})$, and $R_2(\mathbf{S})$. The minimization procedure of the model (12), based on the ADMM algorithm, is detailed in the Appendix.

## IV. Experimental Results

In this section, we focus on impartially describing the performance of the proposed model, since few researchers have focused on oblique stripe removal in remote sensing images, and few of the developed methods can ensure a satisfactory result. To verify the effectiveness of the proposed model in both the procedures of oblique stripe orientation and removal, a large number of tests, including both simulated and real data experiments, were undertaken. For a fair quantitative evaluation of the different experimental data, all the test images were normalized between [0, 1].

### *A. Orientation Experiments*

As the first step in the proposed method, the robustness and stability of the stripe orientation are very important. In order to test the accuracy of the proposed orientation approach, image rotations were repeatedly performed to simulate oblique stripes with certain directions. Six images in different striping cases were first chosen from the remote sensing data (without geo-rectification) as base images. Each base image (with either horizontal or vertical stripes) was then further simulated as an image group by rotating 10 times with different randomly selected angles to enrich the test directions of the oblique stripes. Among them, the base images using in the image group 2 and 3 were separately extracted from HYDICE and Hyperion data, while the rest four ones were all extracted from Terra MODIS data.

TABLE I
PERFORMANCE OF THE PROPOSED METHOD IN THE SIMULATED ORIENTATION EXPERIMENTS (DEGREES)

| Group no. | Min. AE | Max. AE | Mean AE | Std. Dev. of AE |
|---|---|---|---|---|
| 1 | 0.01 | 0.33 | 0.13 | 0.10 |
| 2 | 0.00 | 0.70 | 0.32 | 0.23 |
| 3 | 0.04 | 0.48 | 0.23 | 0.14 |
| 4 | 0.00 | 0.13 | 0.06 | 0.05 |
| 5 | 0.02 | 0.31 | 0.10 | 0.11 |
| 6 | 0.00 | 0.21 | 0.09 | 0.07 |

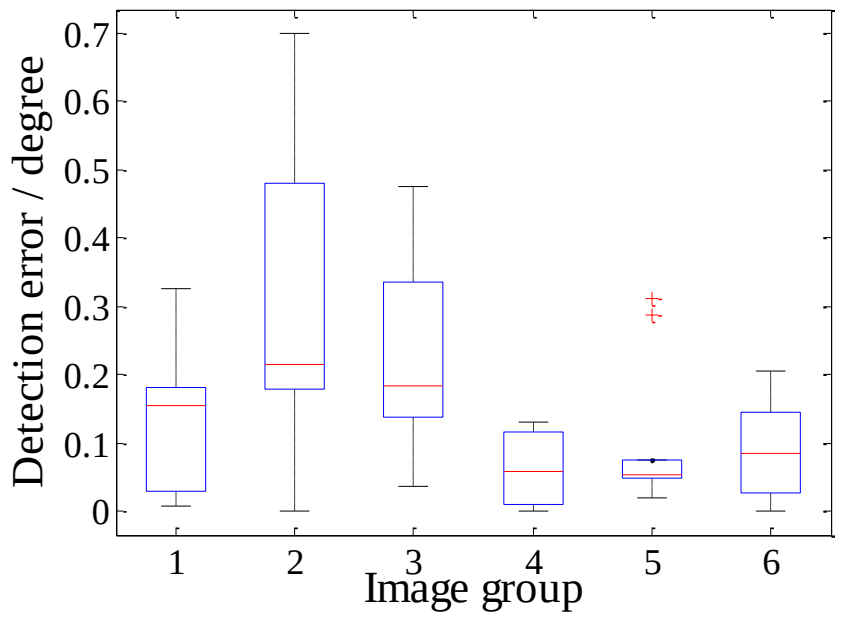

Fig. 6. Boxplots of the orientation results (AE) in the six image groups.

Table I lists the orientation results of the proposed method. For a clear comparison between the different image groups, a more intuitive statistics-based boxplot method [38] was applied and is displayed in Fig. 6. According to the statistics (minimum, maximum, mean, standard deviation) of the absolute errors (AEs) between the simulated and detected oblique stripe orientations in all 60 tests (Table I), the performance of the proposed model is very stable and pleasing. Although, in Fig. 6, the overall errors for the different image groups are different, it can be explicable through a visual analysis. We choose two images with the minimum and maximum orientation AEs from each group as examples, and exhibit them in Fig. 7. For the image groups 1, 4, 5, and 6, the stripes within are distinguishable and possess clear "stripe edges". However, due to the slighter stripe degradation in group 2 and some mixed Gaussian noise in group 3, the "stripe edges" in these two cases are much vaguer and result in relatively higher detection errors as shown in Fig. 6. Fortunately, even when taking these two image groups into account, the highest error as a value of 0.7° is still very slight, which demonstrates the effectiveness of the proposed orientation model.

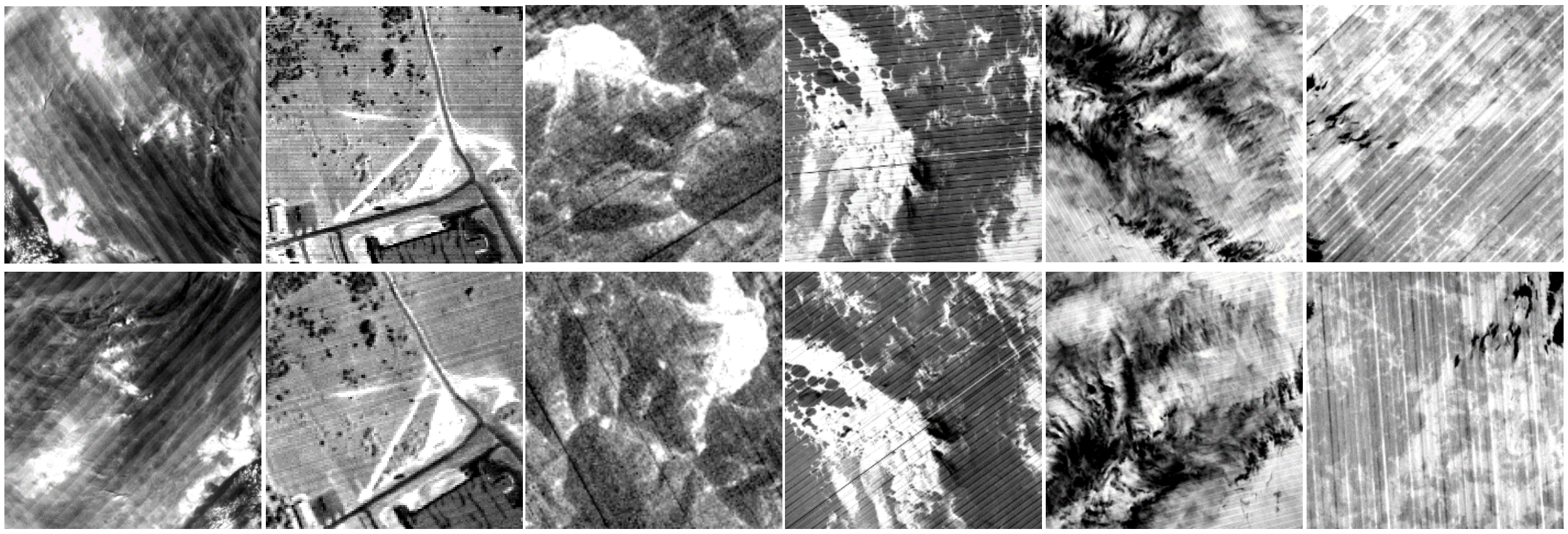

(a) (b) (c) (d) (e) (f)

Fig. 7. Test images for stripe orientation in: (a) image group 1 with 34° and 136° stripes; (b) image group 2 with 90° and 106° stripes; (c) image group 3 with 121° and 29° stripes; (d) image group 4 with 95° and 126° stripes; (e) image group 5 with 30° and 104° stripes; and (f) image group 6 with 135° and 4° stripes. Image rows from top to bottom are images with minimum and maximum orientation AEs.

*B. Destriping Experiments*

In the destriping experiments, two groups of simulated images and four real data images were utilized to test the performance of the proposed method. For the simulations, two clean observations were first degraded by two kinds of striping cases as base images. Specifically, one Terra MODIS image extracted from a heterogeneous area had global vertically distributed random stripes added, and the other Aqua MODIS image from a comparatively homogeneous area was degraded by horizontal periodical stripes. A rotation operation was then repeated 10 times on each base image to simulate a group of oblique stripe images with different directions. Considering the inherent symmetry of the designed oriented variation, we kept the simulated directions of the oblique stripes within the range of $0° \le \theta_{stripe} < 45°$. To ensure a better comparison, 10 randomly selected orientations were shared in the simulation process for two image groups. The final image sizes for these two groups were $200 \times 200$ for the Terra MODIS data (group A) and $400 \times 400$ for the Aqua MODIS data (group B). With regard to the real data experiments, four images extracted from different remote sensing platforms were used: 1) a $400 \times 400$ HJ1A/HIS subimage; 2) an $800 \times 800$ Sentinel 2A subimage; 3) a geo-located $800 \times 800$ Terra MODIS subimage; and 4) a $2000 \times 2000$ LANDSAT 5 TM thermal subimage. Since all the real data except for the Terra MODIS subimage contained different extents of random noise, the stability of the proposed model could be verified through this test.

Full-reference indexes and non-reference indexes were separately applied in the simulated and real data experiments to give the quantitative evaluation. The commonly used full-reference indexes of the mean absolute error (MAE), the peak signal-to-noise ratio (PSNR), the structural similarity index (SSIM) [39], [40], and the feature similarity index (FSIM) [41] were employed in the simulations, while the inverse coefficient of variation (ICV) [42], [43] and the mean relative deviation (MRD) [15], [17] were utilized to assess the destriping performance without the help of true reference data. Generally speaking, ICV can reflect the level of remaining stripe noise, and is calculated via dividing the mean value by the standard deviation of the pixels within the window of a certain size. On the contrary, MRD measures the ability to preserve the original healthy information, and is given by averaging the relative error (multiplied by 100%) between the noisy observation and the destriped data. In practice, the ICV and MRD indexes were computed in the 10×10 image samples chosen from the homogeneous striped and sharp noise-free regions. Higher values of the PSNR, SSIM, FSIM and ICV indexes; and lower values of the MAE and MRD indexes denote an image with a better destriping result.

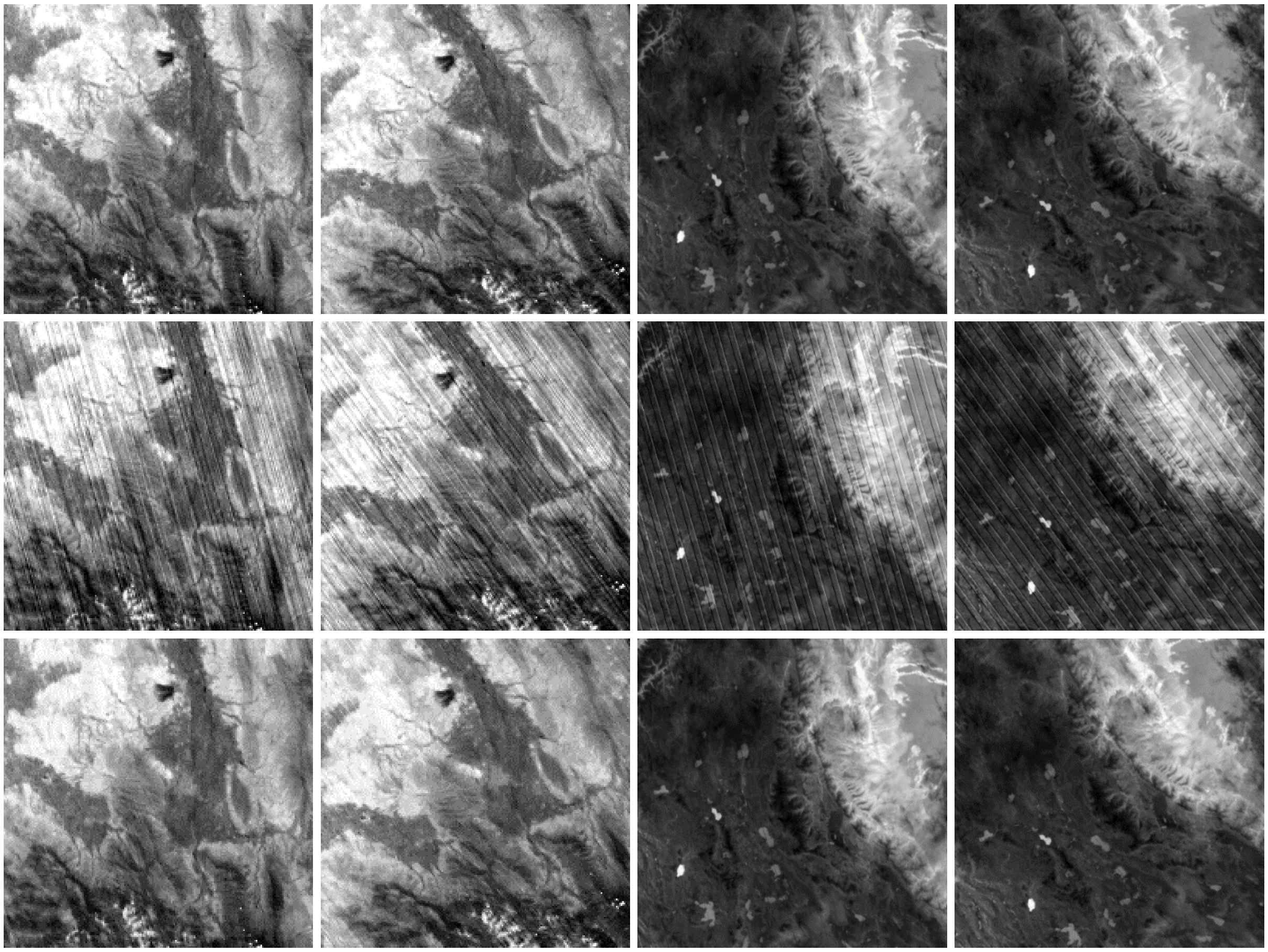

(a) (b) (c) (d)

Fig. 8. Example of the destriping results of the proposed method in the simulated data experiments: (a) the *d*th test in group A; (b) the *h*th test in group A; (c) the *d*th test in group B; and (d) the *h*th test in group B. Image rows from top to bottom are clean observations, degraded images, and the destriping results.

For the simulated experiments, two image groups were exploited to test the proposed method. Among the 20 destriping results, two examples from each group are displayed in Fig. 8. It is clear that both the randomly distributed oblique stripes [Fig. 8(a) and (b)] and the regular periodic oblique stripes [Fig. 8(c) and (d)] are effectively suppressed by the proposed model. Although the detail information in their corresponding clean observations is relatively rich, the visual destriping results without over-processed blurring effects undoubtedly demonstrate a satisfying capability. Moreover, since the base image in each test group remains the same, the most important differences between the simulated experiments within one group lie in the stripe orientations. By comparing the destriping results between Fig. 8(a) and (b) or Fig. 8(c) and (d), the obviously stable performance of the proposed method reflects its advantage of being able to adapt to the change of stripe orientation. More detailed information about the results of the other 16 simulated tests is given quantitatively in Table II.

Similar to the results in the orientation experiments, the differences between the simulated and detected stripe orientations in all the simulated destriping experiments (Table II) are also very small, which once again confirms the accuracy of the orientation step in the proposed method. As for the destriping assessment, a relatively stable performance in MAE, PSNR, SSIM, and FSIM values can be found within one test group for different stripe orientations, which clearly indicates the robustness of the proposed approach in the destriping process.

TABLE II
QUANTITATIVE EVALUATION OF THE FULL-REFERENCE INDEXES IN THE SIMULATED DATA EXPERIMENTS

| Group | No. | Simulated | Detected | MAE(E-2) | PSNR | SSIM | FSIM |
|---|---|---|---|---|---|---|---|
| A | a | 3 | 2.86 | 3.13 | 27.73 | 0.91 | 0.94 |
| | b | 7 | 6.65 | 2.91 | 28.60 | 0.91 | 0.94 |
| | c | 12 | 12.09 | 3.05 | 28.23 | 0.91 | 0.94 |
| | d | 16 | 16.34 | 2.77 | 28.97 | 0.94 | 0.95 |
| | e | 21 | 20.96 | 2.88 | 28.62 | 0.92 | 0.94 |
| | f | 24 | 23.96 | 2.80 | 28.90 | 0.93 | 0.95 |
| | g | 29 | 28.69 | 2.80 | 28.99 | 0.93 | 0.95 |
| | h | 36 | 35.88 | 2.83 | 28.80 | 0.92 | 0.94 |
| | i | 41 | 40.97 | 2.78 | 28.97 | 0.93 | 0.95 |
| | j | 42 | 41.63 | 3.02 | 28.22 | 0.90 | 0.93 |
| B | a | 3 | 2.86 | 0.97 | 34.60 | 0.96 | 0.97 |
| | b | 7 | 6.97 | 0.78 | 37.26 | 0.98 | 0.99 |
| | c | 12 | 11.95 | 0.97 | 35.64 | 0.97 | 0.98 |
| | d | 16 | 15.95 | 0.72 | 37.92 | 0.98 | 0.99 |
| | e | 21 | 20.89 | 0.91 | 36.23 | 0.97 | 0.98 |
| | f | 24 | 24.00 | 0.80 | 37.31 | 0.98 | 0.98 |
| | g | 29 | 29.12 | 0.81 | 37.04 | 0.98 | 0.98 |
| | h | 36 | 36.07 | 0.90 | 36.07 | 0.97 | 0.98 |
| | i | 41 | 40.96 | 0.84 | 36.89 | 0.98 | 0.98 |
| | j | 42 | 42.11 | 0.94 | 35.92 | 0.97 | 0.98 |

We then conducted experiments on the real data, and show the destriping results in Fig. 9. Due to the large image size of the tested LANDSAT 5 TM thermal data, close-ups, for a clearer display, are accordingly given in Fig. 10. It can be observed that the striping cases in these four images are totally different. As shown in Fig. 9(a), the noisy image selected from the HJ1A/HIS data band 51 is highly contaminated by co-existing random stripes and Gaussian noise. With much less mixed noise, the striping patterns in Fig. 9(b), extracted from Sentinel 2A data band 10, are very vague, which challenges the robustness and practicability of the orientation step in the proposed method. Even though the stripes in Fig. 9(c) possess clear "edges" and a certain periodicity, some partially distributed stripes [2] complicate the real situation. Being the largest test dataset [Fig. 9(d)], the close-ups in Fig. 10(a) and (c) show a distinct striping case, where each single stripe possesses a several-pixel width [2].

TABLE III
QUANTITATIVE EVALUATION OF THE NON-REFERENCE INDEXES IN THE REAL DATA EXPERIMENTS

| Image | ICV1 | | ICV2 | | ICV3 | | MRD1 | MRD2 |
|---|---|---|---|---|---|---|---|---|
| | Noisy | Proposed | Noisy | Proposed | Noisy | Proposed | | |
| HJ1A | 4.04 | 12.11 | 6.03 | 12.00 | 7.37 | 12.39 | -- | -- |
| Sentinel | 4.99 | 8.30 | 4.58 | 8.60 | 4.15 | 9.38 | -- | -- |
| Terra MODIS | 0.98 | 74.57 | 0.81 | 80.91 | 0.86 | 67.63 | 1.45 | 0.50 |
| LANDSAT | 15.26 | 275.33 | 18.40 | 100.03 | 25.11 | 148.54 | 0.73 | 0.58 |

Compared to the original noisy images, the results output from the proposed destriping approach upgrade the data quality significantly. The proposed approach not only gets rid of the stripe noise with different features and orientations, but also keeps the healthy fine details, without distortion. For example, the thin river marked in Fig. 9(b) is successfully retained in the destriping process. In addition, the good fidelity of the proposed model is also reflected by the accurate extracted stripe components without additional background information, which is especially apparent for the Terra MODIS data in Fig. 9(c).

The quantitative evaluation for the real data experiments is displayed in Table III. Clearly, the ICV values before and after destriping display a fundamental increase. For instance, in the case of the Terra MODIS subimage without the disturbance of any random noise, the average increase is almost several dozen times the original ICV values, which essentially embodies the powerful destriping capability of the proposed model. Due to the absence of noise-free regions in the HJ1A/HIS and Sentinel 2A data, the MRD index was incalculable and is not given for these two tests. However, the shown MRD results with low values close to zero in the Terra MODIS and LANDSAT 5 TM thermal data illustrate the strength of the proposed method, as does the maintenance of the original noise-free information.

## V. DISCUSSION

### *A. Limitations of the Existing Destriping Methods for Oblique Stripes*

Generally speaking, the vast majority of the destriping methods are customized for stripes with either a horizontal or vertical direction, and to the best of our knowledge, only one publication has described a direct way to remove oblique stripes without a certain restriction on periodicity [6]. However, since this method does not consider the stripe features in geo-rectified images, its destriping performance is highly limited. Additionally, as mentioned in the previous section, all the common destriping methods seem to be adaptable to oblique stripes through a forward and reverse image rotation by a specific angle. But it should also be noted that the jagged effects generated by the image rotation can damage the stripe pattern and have a severe influence on the subsequent destriping process. In order to further illustrate the effectiveness of the proposed method, a destriping comparison with a direct oblique stripe removal model and three commonly used destriping methods (assisted by image rotation) was conducted. The four comparative methods were: 1) wavelet-based oblique stripe removal (WAOR) [6]; 2) moment matching (MM) [12]; 3) combined wavelet-Fourier filtering (WAFT) [7]; and 4) the striping sparsity considered model (SSC) [17]. By taking the visually displayed simulations in the previous destriping experiments as examples, four simulated images were utilized in this test.

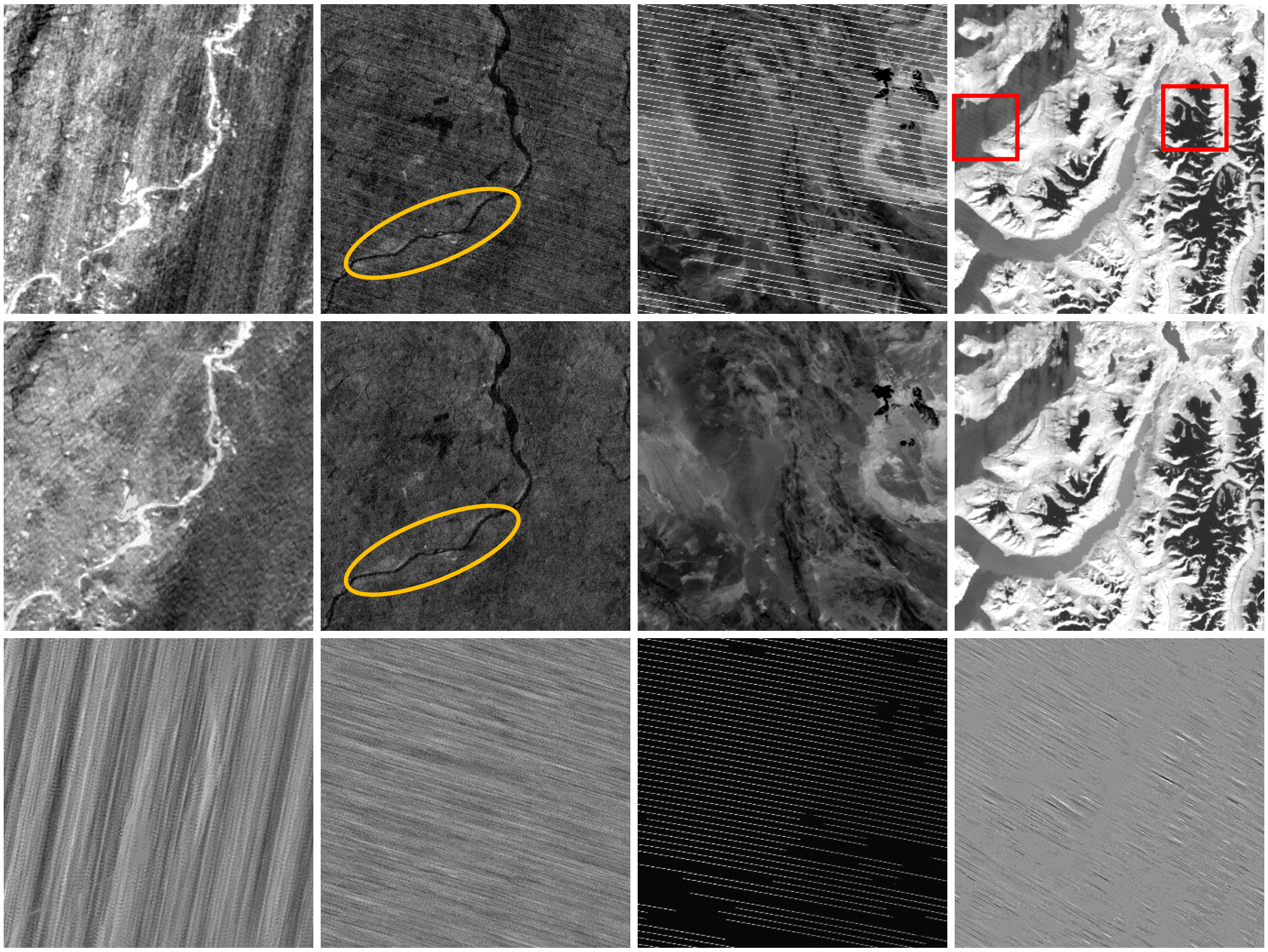

(a) (b) (c) (d)

Fig. 9. Destriping results of the proposed method in the real data experiments: (a) HJ1A/HIS data; (b) Sentinel 2A data; (c) Terra MODIS data; and (d) the LANDSAT 5 TM thermal data. Image rows from top to bottom are original images, destriping results, and the stripe components.

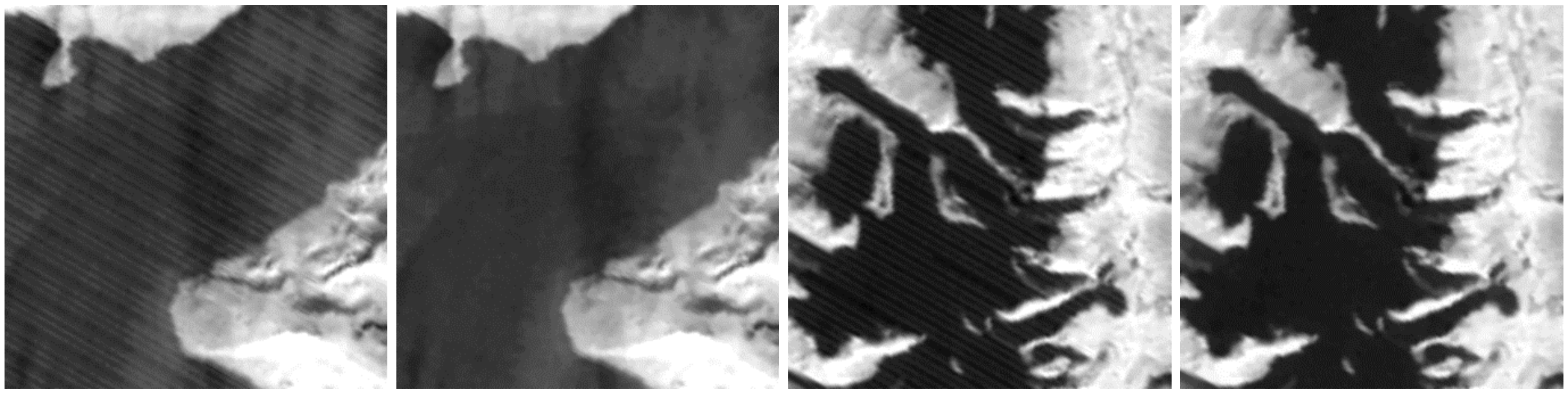

(a) (b) (c) (d)

Fig. 10. Close-ups of the two marked regions in Fig. 9(d): (a) and (c) are from the original image, and (b) and (d) are from the destriping result.

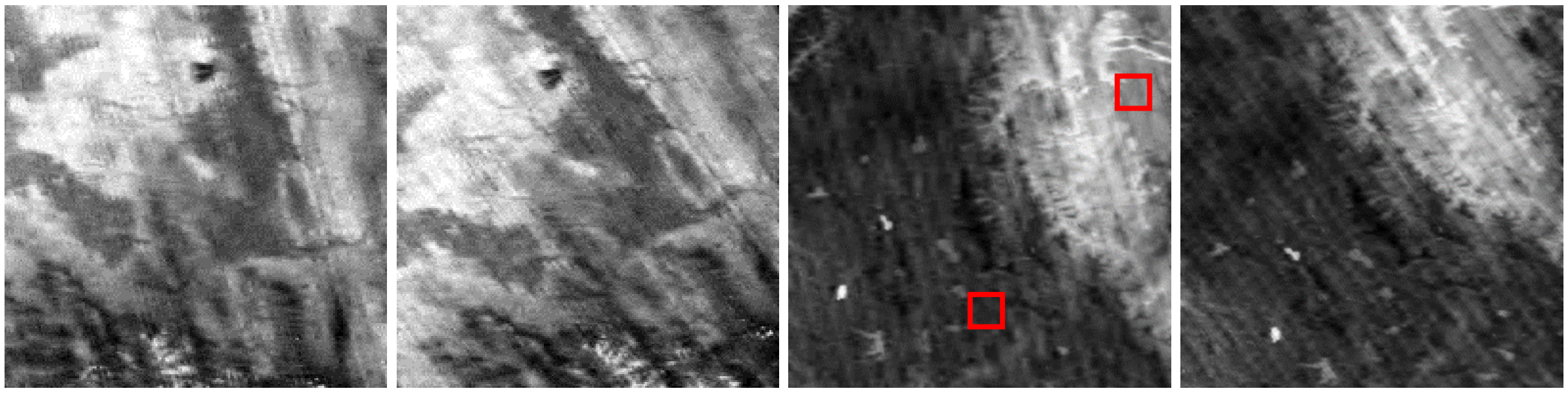

(a) (b) (c) (d)

Fig. 11. Destriping results of WAOR using the data in the simulated destriping experiments: (a) the *d*th image in group A; (b) the *h*th image in group A; (c) the *d*th image in group B; and (d) the *h*th image in group B.

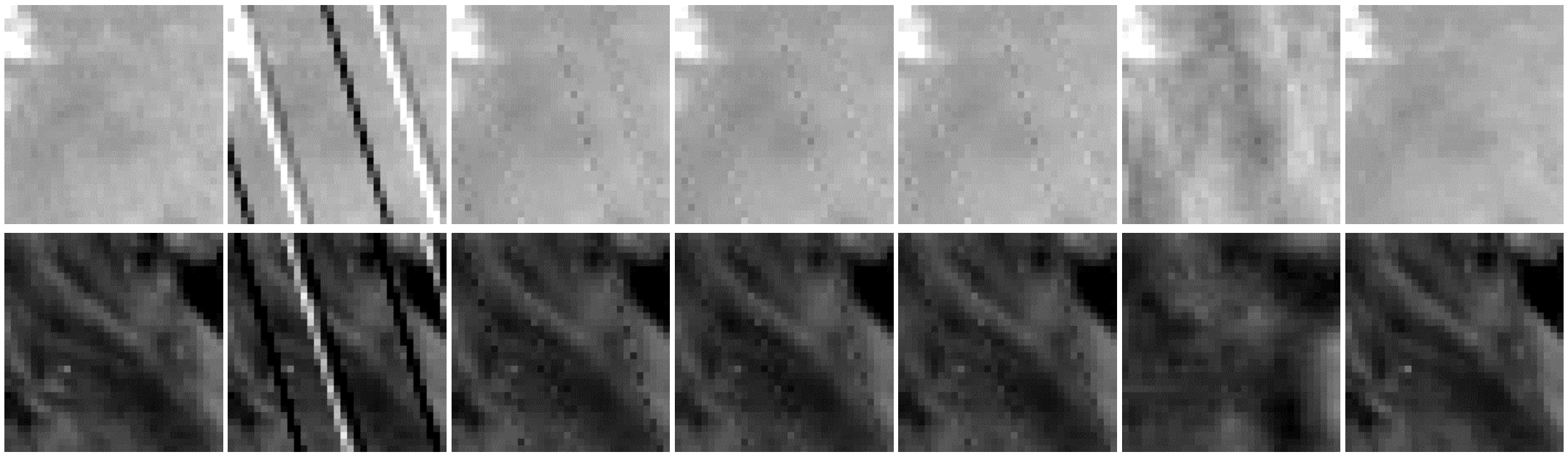

(a) (b) (c) (d) (e) (f) (g)

Fig. 12. Comparison of the close-ups in the two marked regions of Fig. 11(c): (a) clean observation; (b) degraded image; (c) MM; (d) WAFT; (e) SSC; (f) WAOR; and (g) the proposed method.

Fig. 11 shows the destriping results of WAOR. As a method specially designed for oblique stripes, WAOR performs poorly since it neither thoroughly eliminates the stripe noise nor well protects the original healthy information. Its problems with both residual noise and blurred effects are significant in Fig. 11. With regard to the resampling error caused by the image rotation, close-ups from the *d*th image of group B are displayed in Fig. 12 for a clear observation. Among all the destriping methods, the image blurring of WAOR is severe, while the results yielded by the proposed model are the best as they are the most similar to the reference data in Fig. 12(a). The calculated images of the other three commonly used destriping techniques (combined with an image rotation process), present some analogous artifacts such as the regularly distributed dot noise along the oblique stripe orientation [see Fig. 12(c)–(e)]. This is because the jagged stripes after rotation do not conform to the basic assumption of stripe noise, and they cannot be well suppressed by the common destriping methods.

TABLE IV
QUANTITATIVE EVALUATION USING THE FULL-REFERENCE INDEXES IN THE DESTRIPING COMPARISON

| Image | Index | MM | SSC | WAFT | WAOR | Proposed |
|---|---|---|---|---|---|---|
| A-d | MAE(E-2) | 2.77 | 2.82 | 3.13 | 4.51 | **2.77** |
| | PSNR | 28.41 | 28.56 | 28.06 | 24.80 | **28.97** |
| | SSIM | 0.91 | 0.91 | 0.91 | 0.72 | **0.94** |
| | FSIM | 0.94 | 0.94 | 0.94 | 0.86 | **0.95** |
| A-h | MAE(E-2) | 3.10 | 3.26 | 3.10 | 4.41 | **2.83** |
| | PSNR | 27.61 | 27.24 | 27.78 | 25.00 | **28.80** |
| | SSIM | 0.89 | 0.88 | 0.90 | 0.74 | **0.92** |
| | FSIM | 0.93 | 0.92 | 0.93 | 0.87 | **0.94** |
| B-d | MAE(E-2) | 1.21 | 1.06 | 1.27 | 2.91 | **0.72** |
| | PSNR | 35.08 | 35.98 | 35.05 | 28.40 | **37.92** |
| | SSIM | 0.97 | 0.97 | 0.97 | 0.80 | **0.98** |
| | FSIM | 0.98 | 0.98 | 0.98 | 0.91 | **0.99** |
| B-h | MAE(E-2) | 1.47 | 1.31 | 1.51 | 2.99 | **0.90** |
| | PSNR | 33.55 | 34.38 | 33.53 | 28.15 | **36.07** |
| | SSIM | 0.96 | 0.96 | 0.95 | 0.79 | **0.97** |
| | FSIM | 0.97 | 0.97 | 0.97 | 0.90 | **0.98** |

The overall quantitative assessment is listed in Table IV. In accordance with the visual evaluation, the quantitative performance of the proposed approach beats all the comparative methods in terms of all the full-reference indexes, which forcefully demonstrates the significant advantages of the proposed method in oblique stripe elimination.

## *B. Parameter Selection*

Due to the use of oriented variation, the proposed method can accurately model the along-stripe smoothness of oblique stripes. However if the orientation difference between the real stripes and the chosen candidate is too large, an inaccurate direction description in the destriping model (12) will result in an inaccurate output. Therefore, an appropriate choice of template size is the key to guaranteeing sufficient candidate directions in the oriented variation for a stripe-orientation-invariant (stable) destriping performance. To allow us to give a generally valid template radius ($r$), a test was conducted to analyze the influence of the choice of different value of $r$ on the average destriping performance for a set of images with different stripe orientations. The PSNR index was used to quantify the destriping performance. Since the regularization parameters $\lambda_1$ and $\lambda_2$ in model (12) also impact the destriping results, to eliminate or control this factor, the max PSNR value calculated from the whole parameter space composed by $\lambda_1$ and $\lambda_2$ was selected to represent the final destriping performance of a certain image using a given value of $r$. By exploiting two simulated image groups from the destriping experiments as test data, the average destriping performance varying with different template radius values ($r \in [1, 15]$) was computed, respectively, for these two groups, and the results are accordingly shown in Fig. 13(a) for group A and Fig. 13(b) for group B. As a supplement, the change of the theoretically largest orientation difference between the real stripes and the chosen candidate with $r$ is also given in Fig. 13(c).

Slightly different from the case with a clear best average destriping performance occurring at $r = 9$ in Fig. 13(a), a relatively pleasing mean-max PSNR value starting from $r = 9$ can be seen, remaining relatively stable to $r = 15$ in Fig. 13(b). Additionally, in Fig. 13(c), a “boundary” of $r = 9$ can also be found since the largest orientation difference is already small at $r = 9$, and only decreases a little with the sequentially increasing $r$. The highly consistent results in Fig. 13 motivated us to fix $r = 9$ in our experiments, and this value is also recommended in real implementations of the proposed method. As for the reason why the average destriping performance does not continuously increase with $r$, this is down to the fact that stripe noise is not composed of constant-value pixels [13], [14], i.e., the difference between pairs of stripe pixels with a fairly great distance is not necessarily small and suitable for minimization.

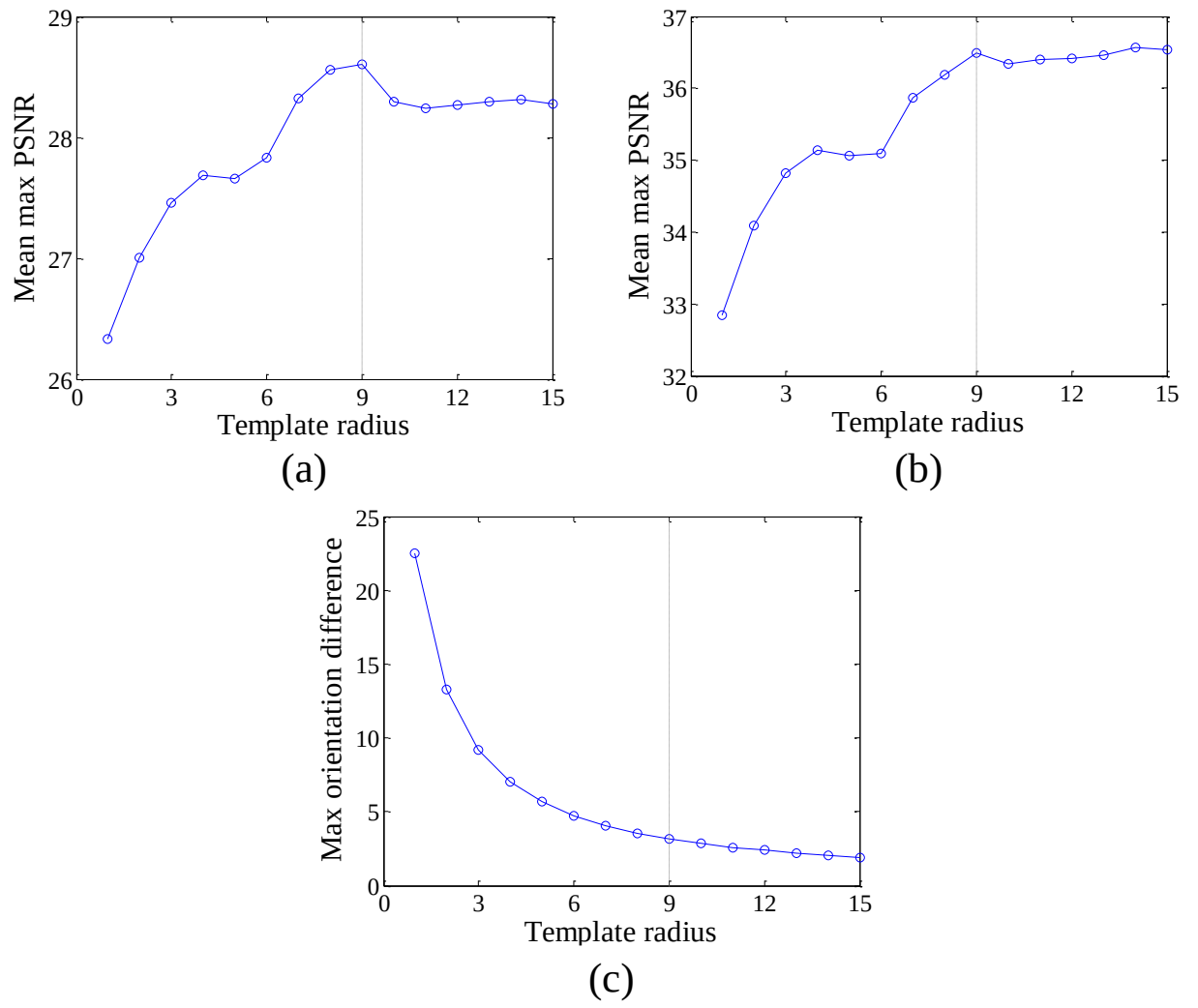

Fig. 13. The performance of the proposed method with different values of template radius $r$: (a) simulated group A; (b) simulated group B; and (c) the theoretically maximum orientation difference between the real stripes and the chosen candidate.

Besides the template radius $r$, the regularization parameters $\lambda_1$ and $\lambda_2$ in the destriping model (12) are also very important. $\lambda_1$ works to characterize the stripe pattern, and the vaguer the stripes are, the larger $\lambda_1$ should be set to ensure a clear separation between $\mathbf{X}$ and $\mathbf{S}$. Meanwhile, to capture the stripe distribution property, a higher value of $\lambda_2$ can be more adaptive to the case with sparsely distributed stripe noise but no co-existing random noise. The effects of $\lambda_1$ and $\lambda_2$ on destriping performance were detailed through parameter sensitivity analyses using PSNR index. By taking the visually displayed simulations in the destriping experiments as examples, the change of PSNR values varying with parameters $\lambda_1$ and $\lambda_2$ is given in Fig. 14. It can be found that the proposed method exhibits satisfying stability with the changes of parameters $\lambda_1$ and $\lambda_2$ both in different degraded stripe cases [when comparing Fig. 14(a) and (b) with (c) and (d)] and in different stripe orientation cases [when comparing Fig. 14(a) and (c) with (b) and (d)]. Therefore, even we have to tune the parameters $\lambda_1$ and $\lambda_2$ in the proposed destriping framework, the tuning process can be easy and convenient. In all of our practical use, the regularization parameters were tuned empirically as $\lambda_1 \in [0.5, 10]$ and $\lambda_2 \in (0, 0.5]$. The selection range of $\lambda_2$ within (0, 0.01] was specifically used for the case with co-existing severe random noise.

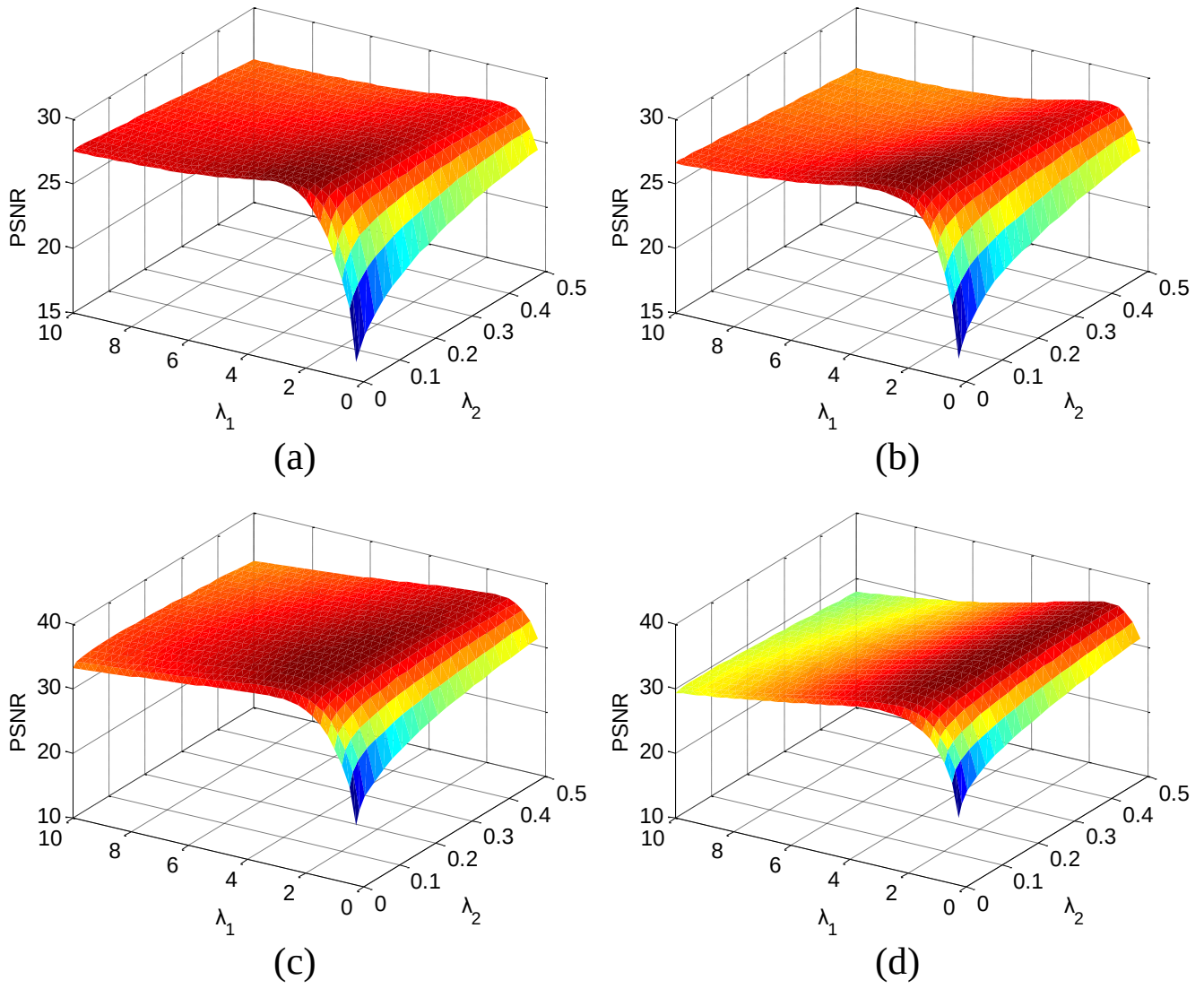

Fig. 14. Sensitivity analysis between parameters $\lambda_1$ and $\lambda_2$ in model (12) using the PSNR index: (a) simulated image A-d; (b) simulated image A-h; (c) simulated image B-d; and (d) simulated image B-h.

Moreover, the two inner parameters in the guided filter for the oblique stripe orientation were fixed as 1 for the window radius and 0.01 for the regularization parameter. The positive factor $t = 5$ was employed in our practical use, which means the background-eliminated images were boosted $\times 5$. The three penalty parameters used in the ADMM algorithm were fixed as $\rho_1 = \rho_2 = \rho_3 = 5$. For the comparative methods, all the parameters were tuned following the authors' recommended rules until obtaining the best destriping results.

### *C. Running Time*

In the proposed method, the orientation for oblique stripes only requires a few operations and can be realized through an efficient guided filter and FFT. Then, in order to estimate the clean image $\mathbf{X}$ from the proposed destriping model (12), the ADMM framework is applied to decompose the whole complicated problem into some simpler sub-problems, see Appendix. The $\mathbf{H}$-related and $\mathbf{V}$-related sub-problems are quickly solved using a soft-shrinkage operator, while the $d$-related sub-problem is calculated through a generalized shrinkage formula. With regard to the $\mathbf{X}$ sub-problem, it has a closed-form solution and can be efficiently computed via FFT. By employing a parallel computation strategy on the $\mathbf{H}$, $\mathbf{V}$, and $d$ sub-problems and the three Lagrange multipliers, we can further reduce the calculation load of the proposed method and make it practicable for large-swath remote sensing images. The running times of the proposed method with different image sizes are given in Table V. It can be seen that the processing time of the proposed approach is quite short at around 1s for a $200 \times 200$ image; and 2 min for a $2000 \times 2000$ image. Since all the experiments in this paper were conducted in MATLAB 2013a on a desktop personal computer with a 3.4-GHz CPU and 8-GB RAM, the proposed model could be accelerated via optimized C.

TABLE V
RUNNING TIMES(S) OF THE PROPOSED METHOD WITH DIFFERENT IMAGE SIZES

| Image size | 200×200 | 400×400 | 800×800 | 2000×2000 |
|---|---|---|---|---|
| Time | 1.18 | 6.55 | 21.41 | 114.85 |

## VI. CONCLUSION

In this paper, we have proposed a novel oriented variation model and constructed a direction-adaptive destriping framework to tackle the challenging oblique stripe removal problem. The oriented variation model is designed to realize the direction approximation of stripe noise after estimating its real orientation from the Fourier domain. Incorporated into the destriping framework, the oriented variation and an $\ell^1$-norm term are functioning to capture the along-stripe smoothness and global distribution property of the stripe noise, while the TV prior is used to describe the feature of the clean image. Experiments in both oblique stripe orientation and suppression were conducted on different test data. The qualitative and quantitative results confirm the excellent effectiveness and stability of the proposed method.

The proposed method can process oblique stripes when their orientation is not altered in the whole image; however for more complicated parallel stripes with varying directions, further in-depth research is still needed. In our future work, we will attempt to deal with this orientation-altered stripe removal problem by fully employing the local information from different image regions.

## APPENDIX

This appendix provides the detailed solution process for the minimization problem of the destriping model (12). Since the degradation model assumes $\mathbf{Y}=\mathbf{X}+\mathbf{S}$, when regarding $\mathbf{X}$ as the only required estimation, (12) can be rewritten as:

$$\hat{\mathbf{X}}=\operatorname{argmin}\{TV(\mathbf{X})+\lambda_1 OV_{\hat{\theta}}(\mathbf{X}-\mathbf{Y})+\lambda_2\|\mathbf{X}-\mathbf{Y}\|_1\} \quad (13)$$

In order to efficiently solve $\mathbf{X}$ from the proposed model (13), the ADMM algorithm is applied in this work to overcome the non-differential and inseparable problem lying in the TV and $\ell^1$-norm term [44]. Based on the ADMM principle, we first introduce four auxiliary variables by setting $d_h=D_h\mathbf{X}$, $d_v=D_v\mathbf{X}$, $\mathbf{V}=D_{\hat{\theta}}\mathbf{X}-D_{\hat{\theta}}\mathbf{Y}$, and $\mathbf{H}=\mathbf{X}-\mathbf{Y}$; thus, (13) can be transformed into the following problem as:

$$\begin{aligned}\{\hat{d}_h,\hat{d}_v,\hat{\mathbf{V}},\hat{\mathbf{H}},\hat{\mathbf{X}}\}=\operatorname{argmin}\{&\|(d_h,d_v)\|_2+\lambda_1\|\mathbf{V}\|_1+\lambda_2\|\mathbf{H}\|_1\\&+\frac{\rho_1}{2}\|D\mathbf{X}-d+\frac{p}{\rho_1}\|_2^2+\frac{\rho_2}{2}\|D_{\hat{\theta}}\mathbf{X}-D_{\hat{\theta}}\mathbf{Y}-\mathbf{V}+\frac{p_2}{\rho_2}\|_2^2\\&+\frac{\rho_3}{2}\|\mathbf{X}-\mathbf{Y}-\mathbf{H}+\frac{p_3}{\rho_3}\|_2^2\}\end{aligned}$$

(14)

where

$$D=\begin{bmatrix}D_h\\D_v\end{bmatrix},\quad d=\begin{bmatrix}d_h\\d_v\end{bmatrix},\quad p=\begin{bmatrix}p_h\\p_v\end{bmatrix} \quad (15)$$

$p_h$, $p_v$, $p_2$, and $p_3$ are Lagrange multipliers, and $\rho_1$, $\rho_2$, and $\rho_3$ are penalty parameters. We then exploit the ADMM algorithm to solve (14), which requires us to minimize the $d$, $\mathbf{V}$, $\mathbf{H}$, and $\mathbf{X}$ sub-problems in an alternating and sequential way.

It is clear that the $\mathbf{V}$-related and $\mathbf{H}$-related sub-problems are well decoupled, so a shrinkage operator [45] can be utilized to explicitly solve these problems as:

$$\mathbf{V}^{k+1}=shrink(D_{\hat{\theta}}\mathbf{X}^k-D_{\hat{\theta}}\mathbf{Y}+\frac{p_2^k}{\rho_2},\frac{\lambda_1}{\rho_2}) \quad (16)$$

$$\mathbf{H}^{k+1}=shrink(\mathbf{X}^k-\mathbf{Y}+\frac{p_3^k}{\rho_3},\frac{\lambda_2}{\rho_3}) \quad (17)$$

where *shrink* is the soft-shrinkage operator:

$$shrink(\alpha,\gamma)=\max(\alpha-\gamma,0)*\frac{\alpha}{|\alpha|} \quad (18)$$

For the $d$-related sub-problem, it reads:

$$\begin{aligned}\{\hat{d}_h,\hat{d}_v\}=\operatorname{argmin}\{&\|(d_h,d_v)\|_2+\frac{\rho_1}{2}\|D_h\mathbf{X}-d_h+\frac{p_h}{\rho_1}\|_2^2\\&+\frac{\rho_1}{2}\|D_v\mathbf{X}-d_v+\frac{p_v}{\rho_1}\|_2^2\}\end{aligned} \quad (19)$$

Despite the fact that the variables $d_h$ and $d_v$ are not totally decoupled, their optimal solutions can still be calculated through a generalized shrinkage formula [46] as:

$$d_h^{k+1}=\max(\mathbf{W}^k-\frac{1}{\rho_1},0)\frac{D_h\mathbf{X}^k+p_h^k/\rho_1}{\mathbf{W}^k} \quad (20)$$

$$d_v^{k+1}=\max(\mathbf{W}^k-\frac{1}{\rho_1},0)\frac{D_v\mathbf{X}^k+p_v^k/\rho_1}{\mathbf{W}^k} \quad (21)$$

where

$$\mathbf{W}^k=\sqrt{|D_h\mathbf{X}^k+p_h^k/\rho_1|^2+|D_v\mathbf{X}^k+p_v^k/\rho_1|^2} \quad (22)$$

Being a typical convex problem, the $\mathbf{X}$-related sub-problem has a closed-form solution via the FFT, which can be formulated as (23), where $\mathcal{F}^{-1}$ represents the inverse FFT.

$$\mathbf{X}^{k+1}=\mathcal{F}^{-1}\left(\frac{\mathcal{F}\left(\rho_1 D^{T}(d-\frac{p^{k}}{\rho_1})+\rho_2 D_{\hat{\theta}}^{T}(D_{\hat{\theta}}\mathbf{Y}+\mathbf{V}-\frac{p_2^{k}}{\rho_2})+\rho_3(\mathbf{Y}+\mathbf{H}-\frac{p_3^{k}}{\rho_3})\right)}{\rho_1(\mathcal{F}(D))^2+\rho_2(\mathcal{F}(D_{\hat{\theta}}))^2+\rho_3}\right) \tag{23}$$

Finally, the Lagrange multipliers $p$, $p_2$, and $p_3$ are updated by:

$$\begin{cases} p^{k+1}=p^{k}+\rho_1(D\mathbf{X}^{k+1}-d^{k+1}) \\ p_2^{k+1}=p_2^{k}+\rho_2(D_{\hat{\theta}}\mathbf{X}^{k+1}-D_{\hat{\theta}}\mathbf{Y}-\mathbf{V}^{k+1}) \\ p_3^{k+1}=p_3^{k}+\rho_3(\mathbf{X}^{k+1}-\mathbf{Y}-\mathbf{H}^{k+1}) \end{cases} \tag{24}$$

The optimization procedure of the proposed destriping model is summarized as Algorithm 1.

**Algorithm 1:** The proposed destriping algorithm

**Input:** data $\mathbf{Y}$, parameters $\lambda_1,\lambda_2,\rho_1,\rho_2,\rho_3$.

**Initialize:** $\mathbf{X}_0=\mathbf{Y},\ p_h=0,\ p_v=0,\ p_2=0,\ p_3=0$, and $\varepsilon=10^{-5}$

**While (** $\|\mathbf{X}^{k}-\mathbf{X}^{k-1}\|/\|\mathbf{X}^{k}\|>\varepsilon$ and $k<N_{max}$ **) do**

Solve $\mathbf{V}^{k+1},\ \mathbf{H}^{k+1}$ using a thresholding method by (16), (17)

Solve $d_h^{k+1},\ d_v^{k+1}$ using a generalized shrinkage by (20), (21)

Solve $\mathbf{X}^{k+1}$ using FFT by (23)

Update the Lagrange multipliers $p^{k+1},\ p_2^{k+1},\ p_3^{k+1}$ by (24)

**End While**

**Output:** $\mathbf{X}^{k+1}$